# Eating Smart: Advancing Health Informatics with the Grounding DINO-based Dietary Assistant App

**Abdelilah Nossair[1], Hamza El Housni[2]**

[1]School of Science and Engineering, Al Akhawayn University, Morocco, a.nossair@aui.ma

[2]School of Science and Engineering, Al Akhawayn University, Morocco, h.elhousni@aui.ma

## ABSTRACT

The Smart Dietary Assistant project combines technology and Machine Learning (ML) to offer personalized advice for people with dietary concerns such as diabetes. This approach focuses on the user helping them make decisions about their diet using the Grounding DINO model. Grounding DINO uses a text encoder and image backbone to improve detection accuracy without relying on a labeled dataset making it practical for real world situations with various food types. This model uses a 52.5 AP score on the COCO dataset and attention mechanisms that leverage features based on user-provided labels and food images to allow precise object recognition. The feature is at the core of the user app, turning smartphones into a helpful dietary advisor that enables people to manage their health effectively.

The app can use your device camera to take photos that will be analyzed by the model for detection and categorize the food items correctly. This is what differs in this system: it decides to be free and not to be connected to annoying cloud databases of information. The application uses a database managed by itself that is of PostgreSQL type, ensuring the preservation of data integrity and control. This database hosting information includes all types of food products, from profiles to health insights drawn from their consumption by human beings. This helps in effective and efficient data access speed, reliability, and enhances user privacy through localized storage within the organizational infrastructure.

The app focuses on improving the experiences of the users, considering that it allows them to create profiles through which they describe themselves based on preferences and tips on nutrition. In addition to calories information, the app provides insights to nutrients such as proteins, vitamins, and minerals. This makes it possible for one to decide the kind of food to take, either for weight management, muscle building, or managing health conditions. On the other part, it also assesses food compatibility versus profiles and gives personal recommendations for alternatives and recipes. Such kind of personal help is highly convenient for persons with needs as it helps them take their healthy options confidently.

Developed using React Native and TypeScript, the Smart Dietary Assistant app guarantees operation across devices and platforms. It incorporates technologies beyond modeling to ensure optimal performance in food recognition, scalability for future enhancements and seamless integration, with other dietary tools.

Users have the option to enjoy features like using the camera to scan food items, for tracking habits and receiving insightful analysis. They can also interact with an assistant for recommendations. The protection of data is ensured through user authentication whereas customizable settings enhance the user experience. React Native enables smooth screen transitions. The expo camera allows scanning capabilities.

Local storage efficiently manages data to create an easy/appealing to use interface.

The Smart Dietary Assistant app's interface stands out for striking a balance between aesthetics and usability. The use of buttons, and a vibrant color scheme enhances user experience by making navigation and feature selection simple. The chatbot feature, represented by an avatar encourages user engagement and personalized guidance seeking. Users find camera scanning convenient although it is noted that varying lighting conditions may affect accuracy. It is this appreciation that opened doors to improvement that can guarantee success in all situations.

The choice of a self-hosted PostgreSQL database for this project re-emphasizes its importance in the realms of health informatics and nutritional science. This is data that can be stored without really depending on outside cloud services, and just with that, the same can be retained as reliable information, since there are chances that it can be changed from the outside. In the future, the Smart Dietary Assistant is planned to be empowered with collaboration with devices. With this development, the application can sync with fitness trackers and smartwatches to give time-based suggestions from physiological data such as blood sugar level and calories burnt. This will connect users to devices that give them individualized advice regarding their health needs, depending on the style of activity. The application is open to collaborations with AI-powered tools in the development of personalized recipes and meal plans that would give the user an easy time adhering to his preferences, dietary restrictions, and time-in sync physiological information. With conditions like diabetes, this holistic approach to diet management is deemed beneficial because it would make the app utilities more effective, always supports objectives for weight management or muscle building, and therefore supports the overall well-being of the user.

Key words: Food Image Recognition, Machine Learning in Nutrition, Zero-Shot Object Detection

# 1. INTRODUCTION

Today's ever-changing health technology landscape now merges with machine learning to open new frontiers of revolutionary health solutions. . The Smart Dietary Assistant application utilizes technology as to give nutritional guidance and be helpful to whom suffers from situation like diabetes[7]. The app uses the 'Grounding DINO' model for object detection to help in improving its capability to recognize and understand more clearly the particular components.

The application should run its core function on the Microsoft COCO dataset, which is richly annotated and has large data support for Computer Vision (CV) tasks[5]. This data helps enhance the app's efficacy in its machine learning algorithms to accurately give dietary recommendations that are both exact and context-relevant. After data preparation and model training, the application will be able to understand food entities in order to give personalized dietary advice.

It has provided safety of users' data through measures like the encryption of data and secure protocols in handling data[9]. This is critical in building users' confidence, more so in line with health data regulations.

Users' feedback collected via surveys have proved the app's performance encompassing the usability, precision in recommendations, and satisfaction by the user[10]. These affirm the positive survey outcomes, which clearly indicate the levels of user content with the application and how well the app meets their needs.

More importantly, the overall approach of the app regarding the user personalized responses, sophisticated data handling ways, and high-level security measures very clearly highlights the app's commitment toward addressing user requirements.

This study investigates the intersection of technology, health information systems, and nutrition science through an analysis of the creation process, functionalities and effects of the Smart Dietary Assistant application [3]. It adds to conversations about the role of health apps in disease management and health promotion while establishing a benchmark for progress in this field [1, 3].

# 2. LITERATURE REVIEW

## 2.1 Research Foundations

### 2.1.1 Microsoft COCO: Common Objects in Context

The paper talks in detail about the Microsoft COCO dataset, designed so that there can be object recognition in scene context. Precise localization of objects in images of real-world scenes is done at the dataset, and thus, it provides instance-level object annotations with object segmentations for the tasks. COCO includes images of objects in their natural world, e.g., supporting context reasoning tasks and those referring to non-standard views.

In this study, the dataset is given the objective of increasing the algorithm's accuracy towards better object detection using the Smart Dietary Assistant app. The annotations will enhance the capability of the app to detect and recognize the instance-level food items in circumstances and environments. This is particularly valuable to the generation of recommendations because the images are showing food items being served unconventionally or otherwise partly hidden. (Lin et al., 2014)

### 2.1.2 A Review on Machine Learning Styles in Computer Vision—Techniques and Future Directions

The research paper thoroughly explores ML methods used in CV. It covers the development and implementation of ML, in CV including supervised reinforcement learning, among approaches. The paper discusses uses such as object recognition, categorization and delves into emerging techniques like Zero Shot Learning (ZSL).
It also looks ahead to prospects and possible enhancements in ML methods for CV emphasizing the significance of innovations like federated learning to boost privacy and data security in distributed settings. This source is valuable for gaining insight into the range of ML applications in CV and the progress aimed at tackling challenges, in the field. "A Review on Machine Learning Styles in Computer Vision—Techniques and Future Directions" (2022)

### 2.1.3 Effectiveness of Mobile Health Interventions on Diabetes and Obesity Treatment and Management: Systematic Review of Systematic Reviews

The study investigates how mobile health (mHealth) interventions impact the management of diabetes and obesity by analyzing systematic reviews and meta-analyses conducted from 2005 to 2019. The findings indicate that mHealth technologies such as applications and text messages have shown effectiveness in enhancing health results like controlling blood sugar levels and managing weight although the level of effectiveness can differ. Some reviews show enhancements in hemoglobin A1c levels and weight loss whereas others observe improvements or no significant advantages. . (Wang et al., 2020)

### 2.1.4 Nutrition-Related Mobile Application for Daily Dietary Self-Monitoring

This paper delves into 205 journal articles spanning from 2007 to 2021, examining the utilization of apps for tracking habits. It underscores an uptick in interest particularly in leveraging mobile applications to enhance dietary choices and overall lifestyle management. The research pinpoints topics like meal tracking and heightened user interaction via mobile health apps. The results indicate that mobile apps serve as resources for evaluating nutrition and steering habits, which aligns directly with the advancement of the Smart Dietary Assistant application. (Ulfa et al., 2022)

### 2.1.5 A study on challenges of data security and data privacy in the healthcare sector: Swot analysis

This article highlights the problems of data privacy and security in healthcare, pointing at the increasing rates of data breaches and cyber-attacks. Despite the progress made in healthcare technology, these vulnerabilities highlight the sector's persistent weaknesses. The SWOT analysis conducted in this research exposes the strengths, weaknesses, opportunities, and threats related to data security in the industry. It also underscores the importance of formulating strategies that help protect sensitive health information. It should be noted that the implementation of robust security measures, such as AES (Advanced Encryption Standard) encryption and secure data protocols is highly recommended. Ensuring these steps are taken is vital for boosting user trust, meeting standards, and safeguarding health information. "A study on Challenges of data security and data privacy in the healthcare sector: SWOT analysis" (2023)

### 2.1.6 Evaluating the User Experience of a Web-based Child Health Record System

The article is a user experience (UX) research of a web-based Child Health Record System applied with a User Experience Questionnaire (UEQ), based on the six-dimension models of UX: Attractiveness, Perspicuity, Efficiency, Dependability, Stimulation, and Novelty. The study shows that these areas are the highly satisfied users; hence, they trust in the importance of a friendly and easy interface to enhance the involvement and contentment of these users. In fact, the study illustrates by real testing how detailed UX of an app will increase its usability. This provides a unique resource for developers looking to increase the design and functionality of health applications. Results display one of the advantageous outcomes of UX design and testing, with substantial improvements in user interaction experiences through health apps. (Wahab et al., 2020)

## 2.2 Gaps in current research

Recent studies of nutrition apps suggest room for improvement in their accuracy of food recognition, personalized dietary advice, and privacy protection of personal data. Most of the solutions that exist heavily rely on the pre-defined datasets, which cannot capture a wide variety of food presented to the users; hence, it is limited to recognizing many types of foods. Most apps also lack sophistication that would enable them to offer tailor-made dietary guidance based on specific dietary needs and preferences. The other big concern is data privacy, since most of the apps are not following the guidelines while protecting and securing user information, it raises question marks on trust and compliance with laws.

The Grounding DINO model attempts to present a solution to the first problematic by incorporating it into a custom architecture. Grounding Dino can zero-shot learn to recognize food items without large labels on them, making the app efficient in handling very diverse dietary inputs. Furthermore, functionalities necessary for personalized recommendations on individual user profiles and dietary restrictions to make the delivered advice relevant and context-aware in terms of practical use were included in the proposed architecture. Moreover, the architecture prioritizes data security by using encryption methods and local data storage to protect user data and address privacy concerns proactively.

These advancements set the Smart Dietary Assistant app apart

as a choice, in the mobile health app market expanding the possibilities of what can be accomplished in dietary management technology.

## 3. THEORETICAL FRAMEWORK

The Smart Dietary Assistant app's theoretical foundation is built on three pillars of thought: the use of ZSL in CV, Personalization Theories in health interventions, and Data Privacy Theories, in health apps.

### 3.1 Zero-Shot Learning in Computer Vision

Our app's image recognition is powered by ZSL—the learning approach mimicking the human method of identifying and categorizing an object based on clues. This requires an understanding of the entities, attributes, and relations between entities to identify the many different instances of food items involved, more especially in cases where detailed labeling may not be possible.

### 3.2. Personalization Theories in Health Interventions

The Smart Dietary Assistant app is developed with theories emphasizing the communication of health. Two theories that propose the effectiveness of tailored interventions in instigating changes in health behavior are the Theory of Planned Behavior and the Transtheoretical Model. The app uses data to customize content in order to enhance user interaction and adherence to advice. This harmonizes with studies showing the rate of success for mobile health applications is high in cases where interventions are flexible.

### 3.3 Data Privacy Theories

In this day and age of healthcare information security, privacy has become paramount. Our approach to safeguard user information mimics the triad in information security and heavily depends on encryption using AES and the Transport Layer Security (TLS) protocols. These principles provide a basis for ensuring that the information belonging to users is not only kept private and secure but also made accessible, conforming to standards and regulations.

## 4. METHODOLOGY

The section of Methodology gives an outline of how the Smart Dietary Assistant application is designed and how data spreads across the sections of it. It may be read as a story of the work of the application and interrelated technologies against the structure of a system.

### 4.1 Introduction to the System Architecture

#### 4.1.1 Authentication

Using Firebase Authentication, the setup ensures every user log-in is safe and secure. Therefore, the setup secures the details and credentials with authentication elements setting up a user journey that is safe right from the start.

#### 4.1.2 Server-Side Framework

The Application Server is built using Python and Django technologies; this support assures reliable back-end processes of the content delivery platform that are highly necessary for the management of intricate user information, since it supports the development of data. This set-up allows for the delivery of content.

#### 4.1.3 Machine Learning Infrastructure

TensorFlow on the Machine Learning Server handles tasks that are crucial for running the Grounding DINO model. This model plays a role in identifying, sorting food items, and improving the app's accuracy in offering suggestions without relying heavily on labeled datasets.

#### 4.1.3 Database Management

PostgreSQL is commonly used for its data integrity capabilities and adherence to SQL standards, which makes it well suited for saving in-depth user profiles and nutritional information about food.

#### 4.1.4 Data Security

Security measures for data involve a two-layer strategy; AES encryption safeguards data stored in the database whereas the TLS protocol secures data during transmission.

### 4.2 Graphical representation of app architecture

This is the architectural representation of the Smart Dietary Assistant app. It shows the flow of data from the user's device to various technological parts used in giving tailored dietary recommendations. Where each of the parts duly functions with the purpose of proofing distinct design of the app, covering user engagement, data handling, ML analysis, security, privacy features.

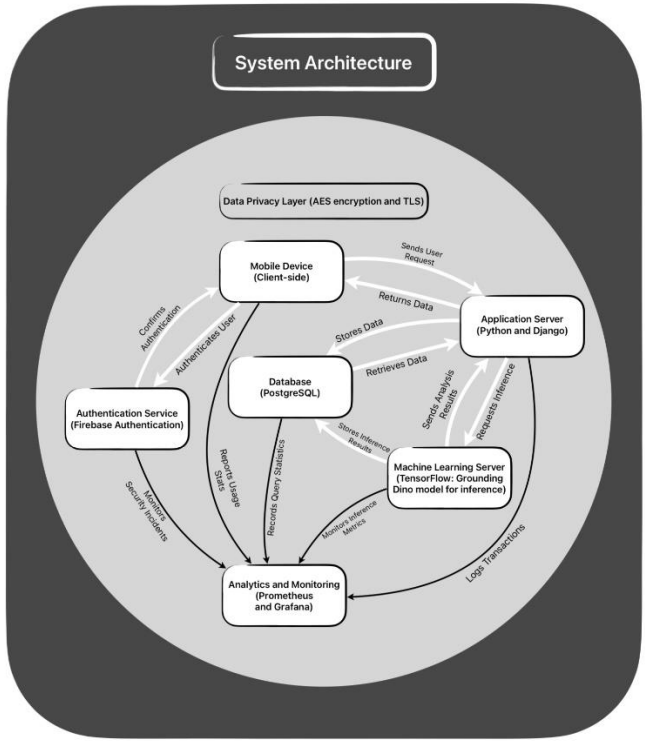

Figure 1: System Architecture of the Smart Dietary Assistant app. This diagram highlights the interconnected nature of the app's components, emphasizing the seamless flow of data and the robust privacy and security layers embedded within the system.

## 4.3 Smart Dietary app data flow process

### 4.3.1 User Interaction

Users begin their journey by signing into the application, which kicks off a session laying the groundwork for a customized engagement.

### 4.3.2 User Profiles and Customization

When users sign up, they share in depth health information, such as diabetes diagnosis. This information plays a role in tailoring the advice offered by the application. The customization algorithms continuously adapt the suggestions recommending foods according to health requirements and glycemic index considerations.

### 4.3.3 Image Capture and Upload

Upon logging in, individuals have the option to take pictures of their meals, which are then securely sent to the Application Server through TLS encryption. This initiates the phase in a series of data processing activities.

### 4.3.4 Backend Processing and Machine Learning Analysis

The Application Server acts as a bridge sending the image to the Machine Learning Server. Here the Grounding DINO model analyzes the image to recognize and classify food items by their traits and acquired patterns.

### 4.3.5 Algorithm Explanation: Grounding DINO in Dietary Management

The system is finely adjusted to evaluate food items according to how they may affect blood sugar levels, which proves useful for managing one's diet as a diabetic. It sorts foods into categories that either help or harm assisting individuals in making informed choices, about their meals.

### 4.3.6 Data Retrieval and Nutritional Analysis

After reviewing the food nutritional details, the specified items are retrieved from the PostgreSQL database, which are organized to provide insights.

### 4.3.7 User Presentation

The organized nutritional guidance, presented in a user-friendly manner appears on the individual's device empowering him/her to make educated choices using the thorough evaluation offered by the application.

## 4.4 Survey Design and Sample Size

### 4.4.1 In-depth Survey Purpose Overview

The Smart Dietary Assistant app conducted an evaluation of user health benefits through a survey. The survey was carefully designed to gather feedback on aspects of the app ranging from how easy it is to use to the accuracy and usefulness of its dietary suggestions, including the speed of its image recognition capability.

### 4.4.2 Elaboration on Survey Question Design

We designed each survey question to gather data that reflects how well the app meets user expectations. For instance, we used a Likert scale to evaluate how easy it is to navigate the app providing insights into user experience. We also examined the accuracy of recommendations to evaluate the effectiveness of the recommendation system. Additionally, we assessed how responsive the app is in identifying food items to ensure efficiency. Questions about advice were included to evaluate how tailored the guidance is for users' needs.

### 4.4.3 Diversity in Response Formats and Implications for Metrics

We used a combination of Likert scales and multiple-choice questions to capture a range of user interactions. The Likert scales ranged from 1 to 5, where 1 represented difficulty or inaccuracy whereas 5 indicated ease or precision. This approach allowed users to assign a tangible measure to their experiences. These ratings directly impacted metrics such as user satisfaction, recommendation accuracy and overall user experience. In addition, the multiple-choice questions explored aspects of system efficiency and user trust in our data privacy and security measures providing insights critical to the credibility and reliability of the app.

### 4.4.4 Rigorous Sample Size Calculation for Representativeness

In line with statistical best practices, we calculated the sample size for the survey by using a formula that considers a 95% confidence level and a margin of error of 5%. This computation assumed having a number of users with a 50% proportion to account for the variation in responses.

$$n = \frac{Z^2 \, p \, (1-p)}{E^2} \quad (1)$$

In the equation, *n* signifies the sample size, *Z* represents the Z-score correlating to the desired confidence level, *p* the assumed proportion, and *E* the margin of error. Based on this calculation, an estimated sample size of 385 people was determined, ensuring that the survey findings accurately represent the viewpoints and encounters of a range of users using the Smart Dietary Assistant app.

### 4.5 Continuous Monitoring and Security

### 4.5.1 Analytics and Continuous Monitoring

In this part of the methodology, we highlight how important Prometheus and Grafana are, in the Smart Dietary Assistant apps ecosystem. Prometheus was used for its monitoring capabilities and the time it took to keep track of the operational metrics to ensure smooth running of the app. It also collects data on the time the server uses to respond, system load, and how frequently features are used. Working in conjunction with Prometheus, Grafana works as the visualization backbone through presentation dashboards that indicate these metrics by use of graphs and charts. The combination of tools in this toolchain allows further maintenance and improvements, ensuring that, by no means, would potential user experience-related problems go undetected or unfixed. Such conscientious analysis and ongoing monitoring contribute to the optimization of the reliability of the app, providing a unique user experience.

### 4.5.2 Ensuring Privacy and Security

In the domain of health, users will trust depending on how well maintained the data security will be. The Smart Dietary Assistant app puts in place a range of security measures. The safeguard works by employing AES as a powerful algorithm to encrypt the database. All personal and sensitive information, as well as preferences and health data, are secured from any outside third-party access. Besides AES, the application also leverages the TLS protocol in securing all the data when in transit from the user's device to the app servers. The cryptographic protocol ensures data integrity and privacy not only from interceptions but will also ensure increased user confidence that the application is dedicated to focusing on security. The application also sets up a level of data privacy conforming to the practices of cybersecurity in the industry through a combination of AES and TLS.

## 5. DATA

### 5.1 Overview

The Smart Dietary Assistant app utilizes the range of images and annotations found in the COCO dataset. By highlighting a variety of food items, this dataset offers the detailed information required for sophisticated ML algorithms.

### 5.2 Preprocessing for Model Compatibility

In our research, we automated the preprocessing of COCO dataset images to set the stage for the Grounding DINO model's image recognition capabilities. To standard 512x512 pixels, the images were resized with the help of Python scripts along with Python Imaging Library (PIL) and OpenCV. Thus, this resolution was chosen for the balance of efficiency and maintaining the image details necessary to recognize and categorize various types of food items precisely by the model.

To make the images even more adapted for model training and ensure uniformity of the input, we used NumPy for color normalization—that is, changing values in the images based on the mean and standard deviation of the dataset. This helped in neutralizing the effect caused by the variation in lighting conditions by making the distribution across all images uniform.

Data augmentation ensured that the model was well-exposed to the natural variability it would face in real life. We utilized the 'tensorflow.image' module, in which we added on-the-fly image manipulations such as cropping to simulate different object positions, horizontal flipping for consistent orientation, and color adjustments to simulate changes in the intensity of illumination and hue shift, besides many other preprocessing pipeline image manipulations.

These methods of fit enhancement were crucial not only for enhancing the fit adaptability of the model in respect to visualizations but also for increasing the fit capacity of the model in relation to applying the knowledge from the dataset in real-life cases.

## 5.3 Data Splitting

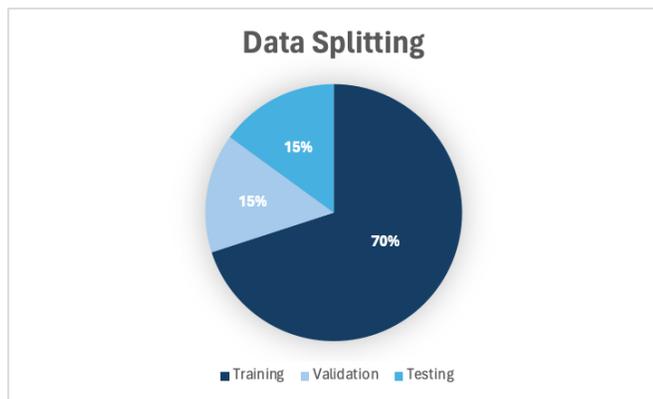

*Figure 2: Data Splitting for Model Training, Validation, and Testing*

The pie chart shows how the data was divided during the model development process. The dataset was split into three parts: 70% for training, and 15% each for validation and testing. This division aimed to improve the model's learning and ability to generalize by having a training set along with portions for validation and testing. The training set is used to adjust the model's parameters, whereas the validation set aids in tuning hyperparameters and avoiding overfitting. The testing set is essential for evaluating how well the model performs on data indicating its effectiveness in real world scenarios.

## 5.4 Dataset Characteristics and Annotations

The detailed representation of scenes and specific annotations, especially focusing on food items, in the COCO dataset is extremely valuable for our nutrition-oriented application. The precise localization of objects through rich instance segmentation is essential for recognizing and examining meal components within settings. This level of detail greatly assists our model's deep learning algorithms in providing accurate analyses.

## 5.5 Data Cleaning and Validation

In order to adapt the COCO dataset to our own needs, we checked the existing annotations and the dataset's categorization and segmentation very carefully. This research aimed to confirm the relevance and correctness of the annotations and labels provided by the workers of the crowd-labor platform Amazon Mechanical Turk[5]. This process ensured that the dataset aligns well with the specific requirements of our food detection tasks, enhancing its applicability and ensuring the reliability of the Grounding DINO model in our application.

## 5.6 Full Dataset Utilization in Model Training

### 5.6.1 Overview

The Smart Dietary Assistant app has been developed in such a way that all the 10,596 food-related images available in the COCO 2017 dataset were used. The size of the dataset and the benefits from effective training determined this decision, necessitated by the need to maintain model neutrality while building.

### 5.6.2 Manageable Dataset Size

The COCO 2017 collection is a total of 10,596 images selected to contain food-relevant content. The size of the dataset is conducive. To ensure balancing between maintaining image details and processing efficiently, the size of the images is reduced to 512x512 pixels. This resolution interfaces rather well with our hardware and makes management of the data very smooth, without downsizing the image. Using the dataset would allow full utilization of our resources and increase our training efficiency without affecting speed or development cost.

### 5.6.3 Comprehensive Training

Training the DINO model with the dataset will let the learning algorithm be made aware of a whole set of variations in food that are present in the dataset. It assures the incorporation of all peculiar food representations existing in cuisines, types of dishes, and cooking techniques while incorporated into the model training process. Rigorous training helps further increase the accuracy of the model and the knowledge applied across food images. It ensures the model not only "adapts" for "learning" from a set of examples but for subtler nuances within the food category.

### 5.6.4 Avoiding Bias

Using the entire dataset opts out for any bias that may come by using just part of the data. Training on all images does not favor categories, whereas at the same time, it avoids the situation of neglecting rarer food categories, which, however, are meaningful. This is crucial in developing an assistant capable of detecting several food items to reflect occurrences in diets as they unfold in reality. This ensures that predictions from the model are unbiased in any food group, food category, or across all the food categories included in the dataset.

# 6. RESULTS

## 6.1 Survey Analysis and User Feedback Interpretation

This part explores the feedback from the 385 users through a survey meant to evaluate performance aspects of the Smart Dietary Assistant app. Here, we analyze the responses to gain insights into user satisfaction and app functionality.

**1. User Friendliness**: With an average rating of 4.20, users perceive the app as intuitive, indicating that the design of the user interface effectively supports easy navigation and accessibility.

**2. Accuracy of Dietary Suggestions:** An average rating of 4.13 suggests that users consider the recommendations relevant showcasing the app's ability to customize dietary advice according to individual health objectives.

**3. Image Recognition Speed:** A satisfaction rate of 65% for 'Very Satisfied' and 'Satisfied' indicates good performance. It also hints at opportunities for improving response times and processing efficiency to enhance user satisfaction.

**4. Personalized Nutritional Guidance**: A rating of 4.04 implies that the app generally caters well to requirements with room for further enhancement to better align with user expectations and preferences.

**5. Trust in Privacy and Security**: The high average score of 4.47 in this aspect showcases confidence in the app's data security measures for maintaining user trust and credibility.

**6. Contentment:** The impressive rating of 4.52 demonstrates a level of user satisfaction highlighting the app's effectiveness in providing a valuable user experience.

**7. Likelihood of Recommendation:** With a rating of 4.38, it suggests that users are highly likely to recommend the app, indicating user satisfaction and the possibility of natural growth.

## 6.2 Net Promoter Score

The Net Promoter Score (NPS), a measure of user support and contentment, is determined through the calculation:

$$NPS = (\% \, Promoters) - (\% \, Detractors) \qquad (2)$$

When ranking from 0 to 5, individuals who rate between 0 - 2 are identified as detractors expressing dissatisfaction. Those giving a score of 3 are considered passives showing neutrality whereas supporters rate the service with a score of 4 – 5, indicating their loyalty and inclination to recommend the service.

An NPS of 41.3 on this scale is favorable illustrating a surplus of promoters compared to detractors. This rating signifies user involvement implying users are inclined to persist in using the service and advocate for it among new users. From a customer experience perspective, an NPS exceeding 40 generally denotes a commitment to the brand or product predicting growth and user retention trends.

## 6.3 Model Performance Metrics

### 6.3.1   Description

Here is an evaluation of how the model performed using a validation set of 1,589 images.

**Correctly Identified Images (True Positives):** The model accurately identified 1,144 images demonstrating its ability to generalize from learned data to new food types effectively.

**Accurately Rejected Images (True Negatives):** It correctly identified 254 images without target food items showcasing its precision in recognizing images that don't contain the specified food items.

**Misclassified Images (False Positives):** There were 116 instances where nonfood items or similar food categories were incorrectly classified, indicating challenges in distinguishing between these categories.

**Missed Food Items (False Negatives):** In 75 images, the model failed to detect food items suggesting areas for improvement in identifying rare foods.

### 6.3.2   Results

The model's precision stands at 90.79% showcasing its accuracy, in predicting images with a proportion of true positives among all positive predictions.

With an accuracy rate of 87.98% the model demonstrates success in classifying data across the dataset.

A recall rate of 93.84% highlights the model's proficiency in identifying actual positive samples.

The F1 Score at 92.30% provides a rounded assessment of precision and recall affirming the model's reliability.

These measurements emphasize the model's effectiveness in categorizing food items through ZSL displaying accuracy, dependability, and excelling in recognizing a range of food items. Improvements in specificity and minimizing positives could enhance performance further.

# 7. Conclusion

The Smart Diet Assistant app is a blend of cutting-edge technology and mobile convenience designed to meet the dietary needs of users, especially those dealing with conditions like diabetes. By utilizing the Grounding DINO model's learning capabilities, it can easily categorize a wide range of food items, effectively handling the complexities of various real life dietary situations.

Feedback from users collected through organized surveys highlights the user interface, accurate dietary recommendations, and overall satisfactory performance of the app. This positive feedback is reflected in an NPS of 41.3 indicating high user satisfaction and a likelihood to recommend the app to other potential users. This strong reception suggests market potential and opportunities for growth.

Looking ahead, the app is set to evolve by integrating with devices. This forward-thinking strategy aims to tailor recommendations more precisely based on real time physiological data aligning closely with individual health needs. These looking advancements are expected to reinforce the app's leadership in healthcare by emphasizing innovation, user involvement, and health information progress.

# 8. Limitations of research

The challenges faced by this project reflect the balance between progress and practical use. The variety of food items in the COCO dataset may not cover all preferences, which could make it challenging for the app's suggestions to be universally applicable. Even though ZSL is a cutting-edge approach, it might not yet match the accuracy of learning methods especially when distinguishing between food items. Moreover, using ML models like Grounding DINO comes with requirements that need to consider device capabilities and resource availability for widespread accessibility. These aspects highlight the need for improvements and adjustments to ensure that the app remains accurate and inclusive across user settings and dietary needs.